# Self-Organizing Multilayered Neural Networks of Optimal Complexity

V.G. Schetinin

Penza State University
40, Krasnaya, Penza, Russia, 440600
vschetinin@mail.ru

**Abstract.** The principles of self-organizing the neural networks of optimal complexity is considered under the unrepresentative learning set. The method of self-organizing the multi-layered neural networks is offered and used to train the logical neural networks which were applied to the medical diagnostics.

## 1 Introduction

The well-known principles of self-organization are used to synthesize the neural networks on the unrepresentative learning sets and eliminate a priori uncertainty into their structures [1]. The self-organization can be realized under next conditions: firstly, if the various structures of the neural network can be generated and, secondly, if the best of them can be selected by a criterion of their efficiency. The variety or the number of the training neural network states must be adequate in accordance with the general principle of W. Ashby [2]. The complexity of the learned neural network will be optimal if its variety will be adequate under the minimal number of its nodes and their synaptic connections.

For known F. Rosenblatt's perceptron consisting of the input (sensor), associative and adjustable layers of the nodes, the complexity is not optimal because of the synaptic links between the its layers are randomly, redundantly defined. It is stated that the supplement of the new layers into its structure improves its recognition capability [3].

The redundancy of the neural network structure can be reduced by the random search methods in which are selected such ones that can decrease the value of a lost function. Within these methods, the search of the desired structure consisting of one associative layer with a priori defined number of the neurons is completed when the defined number of the unsuccessful attempts aimed to decrease the value of lost function is achieved [4-6]. These limitations that without introduced cause to that desired neural network structure can be conditionally optimal.

Within the heuristic self-organization methods, the neural network structure is evolutionary one. The complexity of the neural network is incrementally increased with each new layer until the value of the lost function is decreased. In each layer, a variety of the neural network candidate-structures is generated, and the defined number of the best of them is selected. By using the principle of exterior addition in the selection criteria, the lost function has a minimum that points to the desired neural network function [7-9]. However, the complexity of the synthesized neural network can not be optimal because the results of the heuristic self-organization depend on the defined configure of the selection criterion and the freedom of the candidate-structures selection [5, 10]. Below, we analyze the possibility for self-organizing the multi-layered neural network of the optimal complexity with the suggested approach [11-13].

## 2 Criteria of Efficiency

The behavior of the neural network that has the m input variables $x_1, ..., x_m$ and one output y a function $f(x_1, ..., x_m)$ describes. The self-organization of the neural network is made on the unrepresentative learning set composed of the small number n of the independent instances classified as $y^0_i \in \{0, 1\}$, $i = 1, ..., n$, that belong to the $k = 2$ classes of the distinguished states or the objects. Note that in the case $k > 2$, a sequence of the dichotomy classifications may be used.

Within the heuristic self-organization, the desired neural network is described as the F. Rosenblatt scheme that consists of the sensor and associative layers. The synthesis of the associative layers is made with reference function $g(u_1, ..., u_p)$ of the p arguments $u_1, ..., u_p$, typically $p = 2$. The reference function $g()$ can belong to the arbitrary class of the function (e.g., the Golmogorov-Gabor polynomials, the logical functions).

In each layer r, all the variants of the neural network candidate-structures $f_i$ whose value is called as $z_i$ are generated

$$z_i = g(u_1, u_2). \tag{1}$$

One of the possible algorithms of self-organization is the case

$$u_1 = z_j^{(r-1)}, j = 1, ..., F, \tag{2}$$
$$u_2 = x_k, k = 1, ..., m,$$

where F is the freedom of the selecting the candidate-structures for which the criterion CR value in the layer r is minimal. The number of the combination by 2 from m variables in first layer is $L_1 = C_m^2$, in second and next ones is $L_1 = mF$. Typically, the number F ranges from $0.4L_1$ to $L_1$.

The configure of the criterion CR supposes that the learning set to be separated into the several non-conjunctive subset A, B, ..., which are used for self-organizing the neural network. Typically, these subsets have the same length and their number equals two. Within the known approach, the heuristic is realized which supposes that the true function $f^*$ according to the desire neural network do not depend on the subset A or B that has been chose to synthesize one [8, 9].

For evaluating the efficiency of the neural network candidate-structure $f_i(W/I)$ synthesized on the subset $I = A, B$, the instances of whole set $W = A + B$ are used. This heuristic may be formalized as the criteria of unbias $b_u$ and regularity $\Delta$

$$b_u = |f_i(W/A) - f_i(W/B)|, \tag{3}$$
$$\Delta = |f_i(W/A) - Y^0| + |f_i(W/B) - Y^0|, \tag{4}$$

where $Y^0 = (y^0_1, ..., y^0_n)$ is vector of instruction.

These criteria are called exterior, since for evaluating the function $f_i$ efficiency, the exterior instances from the subset $J \neq I = A, B$ are used. By using the similar configure of the criteria, the principle of S. Beer's exterior addition is realized that allowed to eliminate the contradictories the Godel's theorem about incompleteness of the axiomatic systems does condition [7].

The unbias and regularity values depend on the undetermined component caused due to measurement errors of the input variables as well as the influence of the uncontrolled variables. Because, to increase the robustness of the desired neural network, a convolution CR of criteria (3) and (4) are computed

$$CR = \alpha b_u + \beta \Delta, \tag{5}$$

where $\alpha, \beta \geq 0$ are the coefficients the user defined without.

As we can see, the minimum of this criterion corresponds to the best, desired neural network in the layer r. While number r of the layers is increased, the complexity of the neural network candidate-structures is also done, and the value criterion $CR^r$ comes through a minimum pointed to the desired neural network f*. However, due to the above-mentioned undetermined component, the minimum of this criterion can be local. For eliminating the possible bias of the desired neural network structure, the positive variable δ was introduced in the stopping rule

$$CR^{r-1}_{min} \leq CR^{r-1}_{min} + \delta. \qquad (6)$$

If this condition is carried out in layer r*, the algorithm of self-organization is completed, and the neural network whose value of the criterion equals to $CR^{r-1}_{min}$ is assigned as desirable. Note that the several of neural network structure can have the minimal value of the criterion. Within the known method, for final choosing one of the structures, it is used a subsidiary criterion, the new instances, etc. Since the resulted neural network is found under the minimal number of layers r*, the structure of the features is also minimal.

As we can see, for self-organizing the neural network of optimal complexity, the influence of the settings the user without defines has to be excluded. The results of self-organization depend on the settings and variants because of the next reasons.

(1) The user arbitrarily chooses one of the several variants of dividing the learning set into subsets A and B.

(2) The number F of best candidate-structures can not always be maximal due to the large computational expenses, because it has to be chose less.

(3) The efficiency of the criteria convolution depends on the values α and β.

(4) The efficiency of the stopping rule depends on both the values δ and the noise power.

(5) The final choice of best candidate-structure depends on the type of subsidiary criterion.

Below an algorithm suggested to decide the stated task is discussed [10-13].

## 3 Criteria of Self-Organization

For self-organizing the neural network of optimal complexity, we suggest the exterior criteria whose above-mentioned drawbacks were eliminated. These criteria are realized with computing the value μ of an empirical function introduced to evaluate the neural network accuracy loss that occurs on the whole learning set.

Statement 1. Let the values $\mu_i$, $\mu_j$ and $\mu_k$ of the loss are known correspondingly to the neural networks $f_i^r$, $f_j^{r-1}$ and the feature $x_k$ which are used to decide the task of dichotomy classification. Note that function $f_j^0 = x_j$, $j \neq k = 1, ..., m$. Then for selecting the best neural networks generated in the layer r, sufficiently the next condition to compute

$$\mu_i < \min(\mu_j, \mu_k). \qquad (7)$$

This condition is carried out when the structural modifications the reference function $g(f_j^{r-1}, x_k)$ brings into neural network of the layer r are new ones which do not yet belong to the previous neural network $f_j^{r-1}$. By the definition, these modifications are able to form the exterior addition to the neural network $f_j^{r-1}$ the feature $x_k$ brings. Indeed, if the structural modification is no new, that is, the similar modification has been brought into neural network $f_j^{r-1}$, the condition (7) is not carried out. Consequently, for selecting the best neural networks, this condition is sufficient one.

The number r of the layers is increased until the value $\mu_i$ is decreased and while all the combinations that potentially possess the property of the exterior addition will not be depleted. Therefore, with condition (6), the next rule of stopping can be formulated.

Statement 2. Let $L_r$ be the number of the neural networks in layer r the condition (7) is carried out. Then the algorithm of self-organization can be completed in layer r* if one of two next conditions is carried out

$$\mu_i^r = 0, \tag{8}$$

$$L_{r+1} = 0.$$

Obviously, that if second condition to be, i.e. $L_{r+1} = 0$, then value $\mu_i^r > 0$. In this case, the structure of input variables has to be expanded. If the expanding is not possible or does not bring the results, the learning set must be modified.

When the first condition is carried out, then number $L_r$ may be more than 1. In this case, the neural networks $f_1^*, ..., f_L^*$ that have the same efficiency compose the collective. Note that the number $L_r$ of neural networks into this collective is proportioned to the complexity of the decided task [4, 5]. One of the reasons for appearing collective is the learning set has represented the strongly restricted field of the input variable values.

We can modify the algorithm (2) for generating the neural network candidate-structures in according to the rule (8), if the number F will be equaled to $L_{r-1}$.

$$u_1 = z^{r-1}_j, j = 1, ..., L_{r-1}, \tag{9}$$

$$u_2 = x_k, k = 1, ..., m.$$

When $L_r > 0$, the possibility for evaluating the coherence of the collective decision is appeared. For evaluating the coherence, we can introduce the coefficient $\chi$

$$\chi = l_1/L, \tag{10}$$

where $l_1$ is the number of the neural network voted for the taken decision; L is total number of the neurons in layer r.

Obviously, than the value $\chi$ is close to 1, than the power of coherence is more. The value $\chi$ is maximal on the learning instances, and it is less on another input values.

When the coefficient $\chi$ is less than a value $\chi_0$ the user has without defined, the decision is non- plausible. Typically, the value $\chi_0$ is defined more than 0.8. Analyzing the ration of values $\chi$ and $\chi_0$, the efficiency of learning can be evaluated. In case when $\chi < \chi_0$, either the structure of input variables or the learning set has to be reconstructed.

Synthesized neural network can be clearly represented as the training Steinbuch's matrix [14]. Figure 1 shows an example of the logical neural network using the reference function of two Boolean variables.

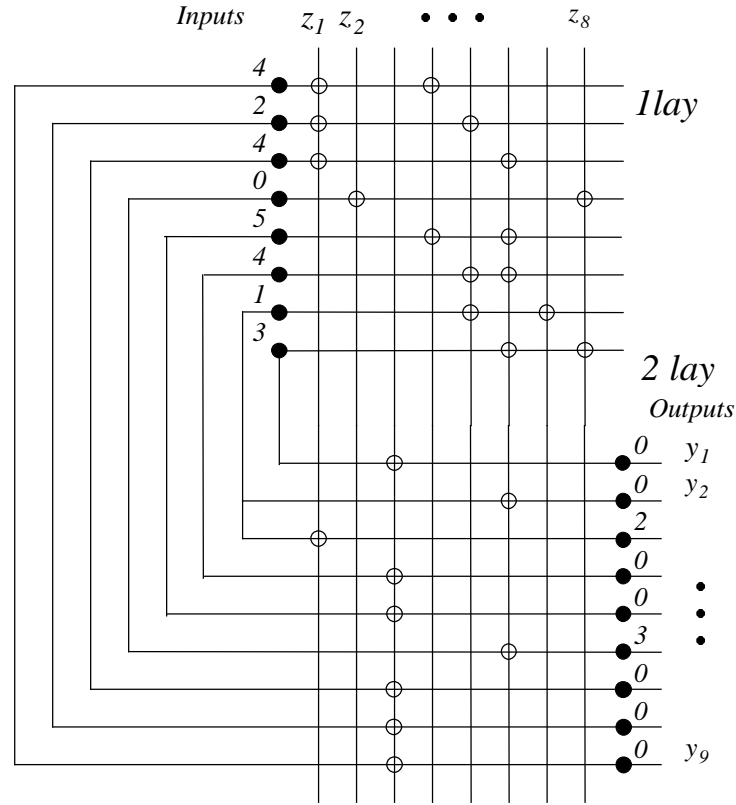

Figure 1: The collective of two-layered logical neural networks depicted as trained matrix including the sensors z1, ..., z8 and the outputs y1, .... y9. The solid circles are the logical functions g0, .., g5 of two variables.

The vertical lines the sensor nodes $z_1, ..., z_8$ fed are Boolean variables. The horizontal lines the hidden nodes fed are described with reference functions $g_i(u_1, u_2)$ depicted as solid circles. The circles placed on the intersection of the vertical and horizontal lines denote that the one of sensor lines to be connected with the input to a formal neuron which implements a reference function $g_i(u_1, u_2)$. Note that number marked near solid circles is the one of 10 logical functions of two Boolean variables. Thus, the horizontal lines are activated under the certain states of the vertical lines.

With the rule (9), only two connections have to be placed on the horizontal lines of the first layer. When the horizontal lines are below intersected with vertical ones, the next layers are formed. Starting at second layer, the rule (9) allows only one connection in the points of intersections. Additionally, the horizontal lines can be split as for second one on Figure 1. Finally, the horizontal lines of second layer do form the variables $y_1, ..., y_9$ which are outputs of the collective consisting of the 9 equal efficiency neural networks.

As we can see, the results of self-organization are the connections between the horizontal and vertical lines as well as the number i of the reference functions $g_i(u_1, u_2)$. Table 1 represents the logical reference functions used to describe the trained neural network.

| Variables | | Function values | | | | | |
|---|---|---|---|---|---|---|---|
| u1 | u2 | g0 | g1 | g2 | g3 | g4 | g5 |
| 0 | 0 | 0 | 0 | 0 | 1 | 1 | 1 |
| 0 | 1 | 0 | 1 | 1 | 0 | 1 | 1 |
| 1 | 0 | 0 | 0 | 1 | 1 | 0 | 1 |
| 1 | 1 | 1 | 0 | 1 | 1 | 1 | 0 |

Table 1: Truth tables for used functions g0, ..., g5 of two variables.

Thus, the trained neural networks can be concisely represented in easily interpreted form.

## 4 Self-Organizing Logical Neural Network

Using the logical reference functions, the neural networks can be synthesized and represented as set of if-then rules. Such neural networks synthesized on unrepresentative learning set are more preferential because their robustness is maximal. Also, for evaluating the decision plausibility, the values $\chi$ of coherence are preliminarily computed for all the $2^m$ combinations of the Boolean features [11-13].

However, for synthesizing the logical neural networks, the quantitative input variables $x_i$ have to be quantized and represented as Boolean or fuzzy ones with the threshold functions

$z_i= 1$, when $x_i \geq u_i$, and 0 if otherwise,

$z_i= 0$, when $x_i \geq u_i$, and 1 if otherwise.

where $u_i$ is threshold for i-th input variable $x_i$.

Note that the threshold value $u_i$ is selected such that the correspondent fuzzy feature $z_i$ should ensure the minimal value of lost function (i.e., the number of the errors) on the learning set.

Using the criterion (7), the so-called problem of the combinatorial "explosion" is avoided. This problem arises if we attempt to search the desired function f* among all the $Q^*= 2^q$ logical functions of the m variables, where $q= 2^m$.

Statement 3. Let $Q(m, r)$ be the maximal number of the logical functions, which can be generated in the layer r using the formulas (1) and (9) with m variables. Then for arbitrary number m variables is carried out inequality

$$\Sigma_r Q(m, r) << Q^*, r= r^*. \qquad (11)$$

In order to prove this, thirst we find the value $Q(m, r)$ under $r= 1$. The number $Q(m, 1)= C^2_m l_0$, where $C^2_m$ is the number of combination by 2 from m; $l_0= 10$ is the number of the logical functions of two variables. Then for $r= 2, 3, ..., r^*$, we find the numbers $Q(m, 2)= mC^2_m l_0^2$, $Q(m, 3)= m^2 C^2_m l_0^3$, etc. Substituted these values in inequality (11), we can see that this inequality is true since the its right-hand increases rapider than its left one. For example, if numbers $m= 5$ and $r^*= 2$ to be, the inequality (11) is carried out because Q equals to 9,940 that substantially less than 4,294,967,296.

## 5 Application of Method

Developed method of self-organization was applied for discovering the diagnostic rules in medicine. The neural network rules were synthesized to distinguish the pathologies that are clinically close each other [11-13]. For example, such pathologies were as (1) the infectious endocarditis (IE) and the active rheumatism (AR), (2) the IE and the systemic lupus erythematosus (SLE), (3) the SLE and the AR. Also, there was (4) the early postoperational complication in abdominal surgery. The diagnostic rules were extracted from the unrepresentative learning sets composed only of the $n= 11 ... 36$ instances the doctors suggested. These instances were represented by the m= 19 ... 31 Boolean and quantitative variables.

To be easily interpreted, the neural network rules were synthesized with the logical reference function of two arguments. The extracted rules were represented as the diagnostic tables or as the set of the logical function if-then. For generating the neural network candidate-structures, the condition (9) was used.

The extracted rules have ensured the unerring classification of the learning sets. These rules include from the 3 up to 8 features whose number is less in 3 times than the doctors suggested. Their accuracy has been tested on the sets contained from the 60 to 100 unseen examples. The first three diagnostic rules are able unerringly to classify their testing sets. The fourth rule generated about 88% true decisions on the 120 unseen examples. In detail, the rule extracted to distinguish the IE and the SLE is below analyzed.

The learning set consists of the n= 36 classified instances which were represented by the m= 31 variables. Among these variables were the 7 quantitative variables. The synthesized neural network, consisting of the 2 layers, the 8 input and 9 output nodes, is depicted on Figure 1 as the training matrix. The input nodes consist of the 2 quantitative features and the 6 Boolean variables that represented in Table 2.

| № и наименование показателя | Обозначение | Порог $u_i$ | $x_i \geq u_i$ | $x_i < u_i$ |
|---|---|---|---|---|
| 1., $10^9$/litre | z1 | 6.2 | z1= 0 | z1= 1 |
| 2., opt. Unit. | z2 | 130 | z2= 0 | z2= 1 |
| 3. | z3 | | | |
| 4. | z4 | | | |
| 5. | z5 | | | |
| 6. | z6 | | | |
| 7. | z7 | | | |
| 8. | z8 | | | |

Table 2: The quantitative and Boolean features that the trained logical neural network uses to differential diagnostics of the IE and the SLE.

In this Table, the thresholds for quantitative features and their functions of quantization are also brought. As we can see, the extracted rule can be represented as the truth table consisted of the $2^8$ rows. For each of the rows, the values of the coefficient $\chi$ can be computed to evaluate the plausibility of the taken decision.

The trained neural network depicted on Figure 1 can be also represented as the set of the logical if-then rules. One of them selected under coefficient $\chi= 6/9$ has the next form

If
    $z_1$= 1 (the leukocytes is less 6.2) and
    $z_2$= 1 (the circulatory immune complex is less 130) and
    $z_3$= 0 (no joint syndrome) and
    $z_4$= 0 (no short wind) and
    $z_5$= 0 (no erythema of skin) and
    $z_6$= 0 (no cardiac noise) and
    $z_7$= 0 (no hepatomegaly) and
    $z_8$= 0 (no myocarditis)
Then
    the pathology is the IE under the 6 from the 9 voted experts.

Recall that the maximal value $\chi$ is 9/9 for any instances belong to the learning set. Accordingly, the above row selected from the truth table under $\chi= 6/9$ corresponds to the unseen example.

## 6 Conclusion

We developed the method of self-organization, which is able to synthesize the multi-layered neural networks of optimal complexity on the unrepresentative learning set. The results of self-organization are not depend on the settings the user defined without. The final decision is taken with value of the coefficient that was introduced to evaluate the coherence of the collective consisted of the neural networks, which have the same efficiency. In particular case, the neural network can be logical one. The developed method was successfully applied to medical diagnostics. In general, the suggested method can use to decide the wide class tasks of the knowledge extraction, making decisions, etc.

**References**

1. Foerster, H., and G.W. Zopf (eds), Principles of self-organization. Pergament Press, 1962.
2. Ashby, W.R., An Introduction to Cybernetics. Willey, New York, 1956.
3. Rosenblatt, F., Princeples of Neurodinamics. New York, Spartan Books, 1962.
4. Rastrigin, L.A., Adaptation of complex systems. Zinatne, Riga, 1981, (in rus).
5. Rastrigin, L.A., and Yu.P. Ponomarev, Extrapolation Methods for Design and Control. Mashinostroenie, Moscow, 1986 (in rus).
6. Glaz, A.B., Applying the Principles of Self-Organisation to Synthesize Decision Rules on the Unrepresentative Learning Set. Automatics, Kiev, 1984, 3, pp. 3-12 (in rus).
7. Beer, S., Cybernetics and Management. English Universities Press, 1959.
8. Ivakhnenko, A.G., Group Method of Data Handling - Rival of Method of Stochastic Approximation. Soviet Automatic Control, 1966, 13.
9. Farlow, S. (ed.), Self-organizing Method in Modeling: GMDH Type Algorithms. Statistics, N.Y., 1984.
10. Green, D.G., and R. Reichelt, Statistical Behavior of the GMDH algoritms. Biometrics, 1988, 44, 49-70.
11. Schetinin, V.G., Synthesizing the minimal decision rules with the principle of the exterior addition. Proceedings of Mathematical Methods of the Pattern Recognition, Moscow, 1995 (in rus).
12. Schetinin, V.G., Synthesizing the minimal neural network for diagnosing an illness. Proceedings of Neuroinformatics and its Applications, Krasnojarsk, 1995 (in rus).
13. Schetinin, V.G., and A.W. Kostunin, Self-Organization of Neuron Collective of Optimal Complexity. Procedings of Nonlinear Theory and its Applications (NOLTA' 96), Kochi (Japan), 1996, pp. 245-248.
14. Shteinbuch, K., Automata and Human. Pergamen Press, 1962.